\documentclass[runningheads]{llncs}
\usepackage[T1]{fontenc}
\usepackage{graphicx,verbatim}

\usepackage{amsmath}
\usepackage{amsfonts}

\usepackage{hyperref} 
\usepackage{color}

\begin{document}
\title{The Right Prior for the Right Deformation: Rethinking Continuous Deformable Image Registration}
\titlerunning{Rethinking Continuous Registration}
%

\author{
Hengjie Liu\inst{1}\orcidID{0000-0001-5890-2291} \and
Chushu Shen\inst{2,3}\orcidID{0009-0006-5865-4754} \and
Dan Ruan\inst{3,4}\orcidID{0000-0003-3400-7684} \and
Ke Sheng\inst{1}\orcidID{0000-0002-6696-5409}
}

\authorrunning{H. Liu et al.}
%
\institute{
Department of Radiation Oncology, University of California, San Francisco,
San Francisco, CA, USA\\
\email{\{hengjie.liu,ke.sheng\}@ucsf.edu}
\and
Biomedical Imaging Research Institute, Cedars-Sinai Medical Center,
Los Angeles, CA, USA
\and
Department of Bioengineering, University of California, Los Angeles,
Los Angeles, CA, USA
\and
Department of Radiation Oncology, University of California, Los Angeles,
Los Angeles, CA, USA
}

\maketitle              
\begin{abstract}
Deformable image registration models implicitly encode deformation priors through their parametrization and optimization. In this work, we conduct a validation study on continuous registration methods to examine how these implicit priors affect performance across different registration tasks. Classic B-Spline transformations impose locality, smoothness, and scale through their control-point structure, whereas recent INR-based methods impose different priors through neural parameterization and optimization. We compare INR-Dense (IDIR), which directly models a dense displacement field using a SIREN-based INR; INR-BSCP (SINR), which predicts B-Spline control points with an INR; D-BSCP, which directly optimizes single-scale B-Spline control points; and MR-D-BSCP, which adds a multiresolution coarse-to-fine scheme. Experiments on inter-subject brain MR registration (OASIS) and intra-subject exhale-to-inhale lung CT registration (DIR-LAB 4DCT) reveal different behavior across deformation regimes. On OASIS, where deformations are moderate but locally complex, D-BSCP matches or slightly outperforms INR-BSCP, suggesting that the B-Spline parameterization accounts for much of INR-BSCP's effectiveness. On DIR-LAB 4DCT, where respiratory motion is larger and more coherent, single-scale B-Spline methods (D-BSCP and INR-BSCP) are less suitable, while INR-Dense and MR-D-BSCP are more effective. Across both tasks, MR-D-BSCP achieves the best performance among the tested continuous parameterizations. These findings highlight that registration accuracy depends strongly on matching the induced deformation prior to the target motion pattern, and support prior-deformation matching as a practical design principle for medical image registration. Our code will be available at \url{https://github.com/HengjieLiu/RightPriorDIR}.

\keywords{Continuous Deformable Image Registration \and Implicit Neural Representation \and B-Spline \and Validation.}

\end{abstract}
%
%
%

\section{Introduction}

Deformable image registration (DIR) estimates a nonlinear spatial transformation that aligns corresponding anatomy across images. It is central to various medical image applications including longitudinal analysis, atlas construction, motion correction, image-guided intervention and so on. The transformation can be parameterized either as a continuous function or as a discrete field. Classical B-Spline free-form deformation (FFD) has been a successful continuous method because it reduces the degrees of freedom while flexibly representing diverse nonrigid deformations ~\cite{FFD1999,Elastix2010}. Discrete dense displacement fields, such as those optimized by Demons ~\cite{Demons1998} and MRF-based methods~\cite{MRFLung2013}, are also widely used.

In the past decade, discrete displacement fields parameterized by deep neural networks became popular due to strong performance and fast inference after population training~\cite{VoxelMorph2019,TransMorph2022,Learn2Reg2023}. However, learning-based methods can degrade in out-of-distribution settings, and pair-specific optimization remains the most robust general-purpose option~\cite{FireANTs2026,LUMirage2026}. Implicit neural representations (INRs) have recently renewed interest in continuous pair-specific registration~\cite{IDIR2022,INRBrain2023,SINR2024,Dual-INR2026}. Instead of storing deformation on a voxel grid, an INR represents it as a coordinate-conditioned function, typically a multilayer perceptron (MLP), that maps continuous spatial coordinates to displacement vectors. This enables arbitrary coordinate queries and analytic derivatives, naturally supporting gradient-based regularization. IDIR~\cite{IDIR2022} first demonstrated this idea for exhale-to-inhale lung CT registration using a SIREN MLP~\cite{SIREN2020}. SINR later argued that dense INR displacement prediction can be suboptimal for inter-subject brain MR registration, and instead used an INR to predict B-Spline FFD control-point values~\cite{SINR2024}.

These results raise an often-overlooked question: when continuous registration improves, are the gains due to the INR, the spline parameterization, or a better match between the model and the deformation pattern? This matters because deformation patterns vary substantially across registration tasks. For example, inter-subject brain MR registration often involves moderate displacements with complex local variation, whereas exhale-to-inhale lung CT registration exhibits larger, more directionally coherent, and smoother respiratory motion. Thus, a continuous "off-grid" model effective in one setting may not generalize to another.

\begin{figure}
\centering
\includegraphics[width=\textwidth]{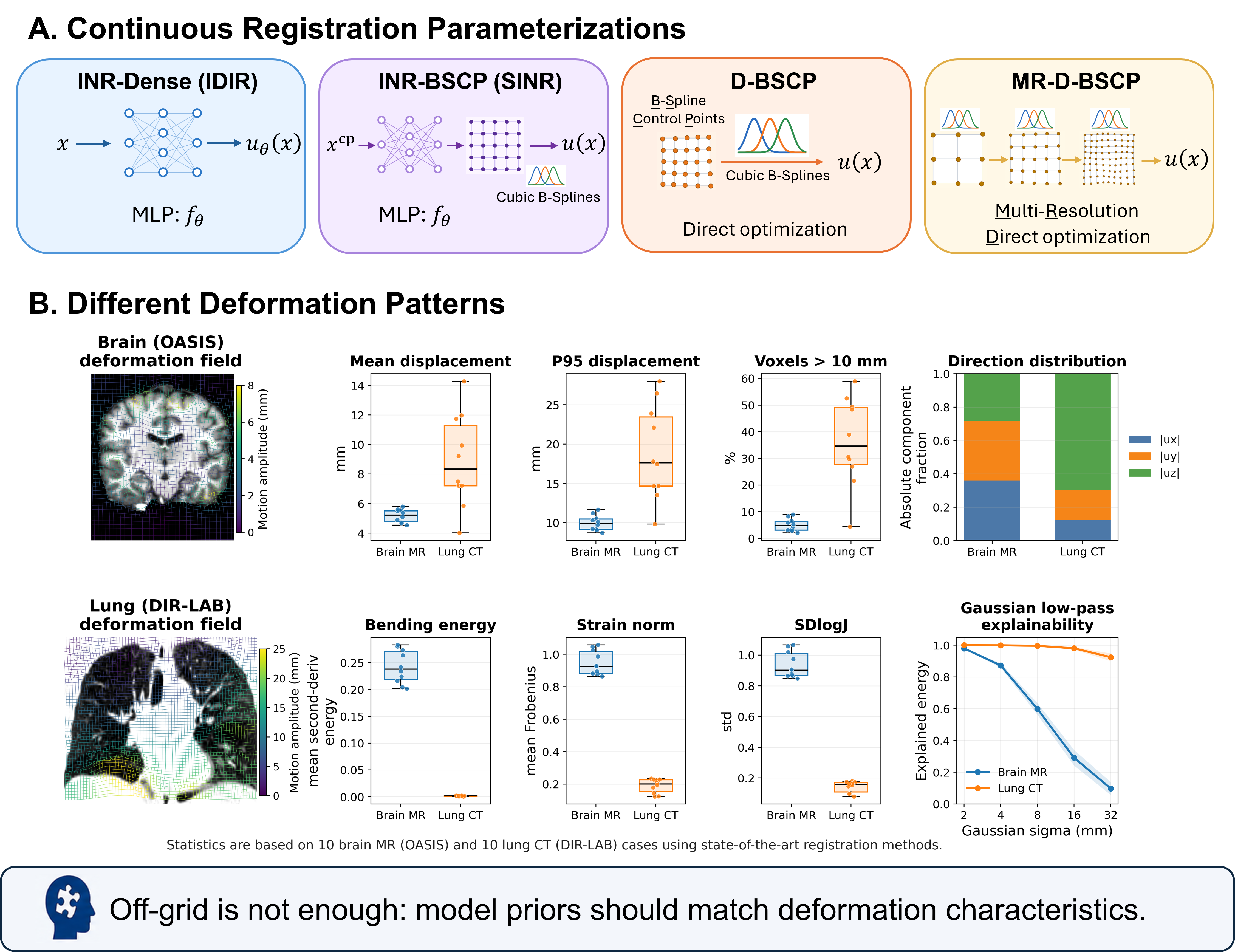}
\caption{Overview of the tested continuous registration parameterizations and registration tasks.}
\label{fig:overview}
\end{figure}

We propose a validation study of continuous DIR parameterizations from this prior-matching perspective, focusing on their accuracy-regularity spectrum~\cite{ARC2025}. For clarity, throughout the remainder of the paper, we use unified notation based on deformation parameterization: \textbf{INR-Dense} for IDIR, which directly models a continuous displacement field with a SIREN-based INR; \textbf{INR-BSCP} for SINR, which predicts B-Spline control points with an INR; \textbf{D-BSCP} for direct B-Spline control-point optimization; and \textbf{MR-D-BSCP} for its multiresolution coarse-to-fine extension. We compare these four core parameterizations on inter-subject brain MR registration and intra-subject exhale-to-inhale lung CT registration (Fig.~\ref{fig:overview}). We additionally evaluate the recently proposed Dual-INR~\cite{Dual-INR2026}, a multiresolution INR approach for lung DIR, on the lung registration task and denote it \textbf{MR-INR-Dense}.

We find that D-BSCP can match INR-BSCP on brain MR registration, suggesting that the B-Spline parameterization explains much of INR-BSCP's performance. For lung CT, however, single-scale B-Spline methods, including INR-BSCP and D-BSCP, are less suitable for large respiratory motion, whereas INR-Dense remains strong and MR-D-BSCP substantially improves the spline-based formulation. MR-INR-Dense also provides a modest improvement over INR-Dense. Overall, MR-D-BSCP achieves the best performance. Our analysis supports a simple but often overlooked principle: registration quality depends strongly on how well the model-induced deformation prior matches the target deformation pattern, including displacement range, smoothness, locality, directionality, frequency content, and optimization path. This also underscores the need for careful validation of new DIR methods to identify which components actually drive registration performance~\cite{RethinkReg2024,RevisitDIR2025}.

\section{Methods}
\subsection{Problem Formulation for DIR}

DIR estimates a displacement field $\mathbf{u}(\mathbf{x})$ that aligns a moving image $M$ to a fixed image $F$ by minimizing the following objective function:
\begin{equation}
\hat{\mathbf{u}} =
\mathop{\mathrm{arg\,min}}_{\mathbf{u}}
\left[
\mathcal{L}_{\mathrm{sim}}(M \circ \boldsymbol{\phi}, F)
+ \lambda \mathcal{R}(\mathbf{u})
\right],
\label{eq:dir_objective}
\end{equation}
where the spatial transformation is given by $\boldsymbol{\phi}(\mathbf{x}) = \mathbf{x} + \mathbf{u}(\mathbf{x})$. Here, $\mathcal{L}_{\mathrm{sim}}$ denotes the image-similarity loss, $\mathcal{R}$ denotes the regularization term, and $\lambda$ controls the trade-off between similarity and regularity. Following the alignment-regularity evaluation framework of~\cite{ARC2025}, we compare methods across regularization strengths to enable fair comparison and identify Pareto fronts.

\subsection{Tested Continuous Registration Models}

Continuous registration methods represent the displacement field
$\mathbf{u}: \Omega \rightarrow \mathbb{R}^{3}$ as a continuous function over the image domain $\Omega$.

\subsubsection{B-Spline FFD.}
It parameterizes $\mathbf{u}(\mathbf{x})$ by a regular lattice of control-point displacements. For a spatial location $\mathbf{x}$, the displacement is computed by tensor-product cubic B-Spline interpolation:
\begin{equation}
\mathbf{u}(\mathbf{x}) =
\sum_{l=0}^{3} \sum_{m=0}^{3} \sum_{n=0}^{3}
B_l(\xi) B_m(\eta) B_n(\zeta)
\mathbf{c}_{i+l,j+m,k+n},
\label{eq:bspline_ffd}
\end{equation}
where $\mathbf{c}_{i+l,j+m,k+n} \in \mathbb{R}^{3}$ denotes a neighboring control-point displacement, $(\xi,\eta,\zeta) \in [0,1]^3$ are the normalized local coordinates of $\mathbf{x}$ within the enclosing lattice cell, and $B_l$, $B_m$, and $B_n$ are cubic B-Spline basis functions. Due to compact support, each displacement depends only on a local control-point neighborhood, and each control point affects only a finite spatial region. Thus, the control-point spacing sets the deformation scale: coarser grids favor smoother, more global deformations, whereas finer grids allow more localized motion.

We evaluate both single-scale direct B-Spline control-point optimization (\textbf{D-BSCP}) and its multiresolution variant (\textbf{MR-D-BSCP}). D-BSCP optimizes control-point displacements at a fixed spacing, e.g., 4 voxels or mm. MR-D-BSCP uses a coarse-to-fine schedule, e.g., 32-16-8-4, with a matching image pyramid in which images are smoothed and downsampled at coarser stages. Each stage initializes the next finer lattice from the previous stage. While general spacing schedules require field upsampling via refitting~\cite{Elastix2010}, our dyadic $2 \times 2 \times 2$ schedule yields nested B-Spline lattices, enabling fast analytic knot insertion.

\subsubsection{INR-Dense (IDIR).}
$\mathbf{u}$ is parameterized with a coordinate-based SIREN MLP, $f_{\theta}$, which takes any spatial coordinate $\mathbf{x} \in \Omega$ as input and directly predicts the displacement:
\begin{equation}
\mathbf{u}_{\theta}(\mathbf{x}) = f_{\theta}(\mathbf{x}), \qquad
\boldsymbol{\phi}_{\theta}(\mathbf{x}) = \mathbf{x} + \mathbf{u}_{\theta}(\mathbf{x}).
\label{eq:idir_dense}
\end{equation}
The network is optimized for each image pair. Because $f_{\theta}$ is differentiable with respect to its input coordinates, spatial derivatives of the transformation can be computed by automatic differentiation.

\subsubsection{INR-BSCP (SINR).}
It uses the same coordinate-based SIREN MLP but predicts B-Spline control-point displacements instead of dense displacements:
\begin{equation}
\mathbf{c}_{p,q,r} = f_{\theta}(\mathbf{x}^{\mathrm{cp}}_{p,q,r}),
\label{eq:sinr_control_points}
\end{equation}
where $\mathbf{x}^{\mathrm{cp}}_{p,q,r}$ is the coordinate of a B-Spline control point. The final displacement field is then obtained by Eq.~\ref{eq:bspline_ffd}.

\section{Experiments}
\subsection{Datasets and Baselines}
We evaluated two registration tasks: inter-subject brain MR registration on OASIS~\cite{dataset-OASIS2007} and intra-subject exhale-to-inhale lung CT registration on DIR-LAB 4DCT~\cite{dataset-DIR-LAB2009}. OASIS contains 414 1mm isotropic T1-weighted MR images. We used the 314/20/80 train/validation/test split for comparison with learning-based baselines and evaluated 100 randomly sampled test pairs. Baselines included four learning-based methods: VoxelMorph~\cite{VoxelMorph2019}, TransMorph~\cite{TransMorph2022}, VFA~\cite{VFA2024}, and SITReg~\cite{SITReg2024}, with VFA and SITReg among the top LUMIR challenge benchmarks~\cite{LUMIR2026}. We also included Greedy~\cite{Greedy2016,Mirage2024} as a strong optimization-based baseline.

For DIR-LAB 4DCT, we evaluated all ten 4DCT cases using the two extreme respiratory phases (T00 and T50). We followed the convention of pTVReg~\cite{pTVReg2017}, a state-of-the-art optimization-based method for this task: scans were resampled to 1mm isotropic spacing for registration, while landmark TRE was computed after resampling the displacement field back to the original image grid.  Since this preprocessing differs from the original INR-Dense setting, we reran INR-Dense on the isotropic images. Given the small cohort of only ten registration pairs, we did not evaluate learning-based baselines on this task.

\subsection{Implementation Details}
We used the official IDIR and SINR codebases to implement INR-Dense and INR-BSCP, respectively, keeping their SIREN architectures, sampling strategies, image-similarity losses, and optimizer settings unchanged. Both use three hidden layers with 256 units per layer. For MR-INR-Dense, we adapted the official Dual-INR implementation and evaluated it on DIR-LAB 4DCT, removing only the lobe-segmentation loss, which we found had negligible effect on performance. We implemented D-BSCP and MR-D-BSCP in PyTorch to directly optimize B-Spline control-point displacements, isolating the effect of the INR from the B-Spline control-point parameterization.

All methods were optimized with Adam for a total budget of 2,500 steps to ensure convergence, with MR-D-BSCP steps divided across resolution levels. Because INR-based and direct B-Spline models have different parameter scales and conditioning, we did not enforce a shared learning rate. Instead, we applied the same coarse sweep over $\{10^{-2},10^{-3},10^{-4},10^{-5}\}$ to both families and selected the best stable setting for each. For INR-based methods, $10^{-2}$ and $10^{-3}$ were unstable, while the official $10^{-4}$ setting was stable and therefore selected. For direct B-Spline methods, we used $10^{-3}$, which was stable and gave the best final performance, while $10^{-2}$ converged faster but slightly reduced accuracy. All experiments were run on NVIDIA RTX 6000 Ada Generation GPUs with 48 GB of memory.

\subsection{Evaluation}
Rather than comparing methods at a single selected hyperparameter, we used the alignment-regularity characteristic (ARC) framework for a fairer comparison across methods~\cite{ARC2025}. ARC compares the full alignment-regularity trade-off spectrum, avoiding conclusions based on a single operating point where better alignment may come at the cost of deformation regularity. For OASIS, alignment was measured by the mean Dice score and mean 95th percentile Hausdorff distance (HD95) over 35 segmentation labels. For DIR-LAB 4DCT, it was measured by landmark target registration error (TRE) over 300 manually annotated landmark pairs per case. Regularity was measured within foreground masks using the standard deviation of the log-Jacobian determinant ($\mathrm{SD}(\log J)$). Fixed external baselines (four learning-based methods and Greedy) were included as single operating points.

We further performed a target-field fitting diagnostic to separate model representation capacity from registration capacity. Each parameterization was trained with a supervised mean-squared error (MSE) loss to fit a strong external reference displacement field: SITReg for OASIS and pTVReg for DIR-LAB 4DCT, both selected from high-performing Pareto-front methods. These fields were not treated as ground truth, but as plausible high-performing deformations for testing how well each parameterization can represent the target when given directly. We ran supervised fitting on 10 random pairs from OASIS and all 10 pairs from DIR-LAB 4DCT, and compared with the unsupervised registration optimization. For unsupervised registration, we selected regularization weights based on the Pareto sweeps so that the compared methods operated at approximately matched deformation regularity, measured by foreground $\mathrm{SD}(\log J)$.

\section{Results}
\subsection{Registration Results}

\begin{figure}
\centering
\includegraphics[width=0.80\textwidth]{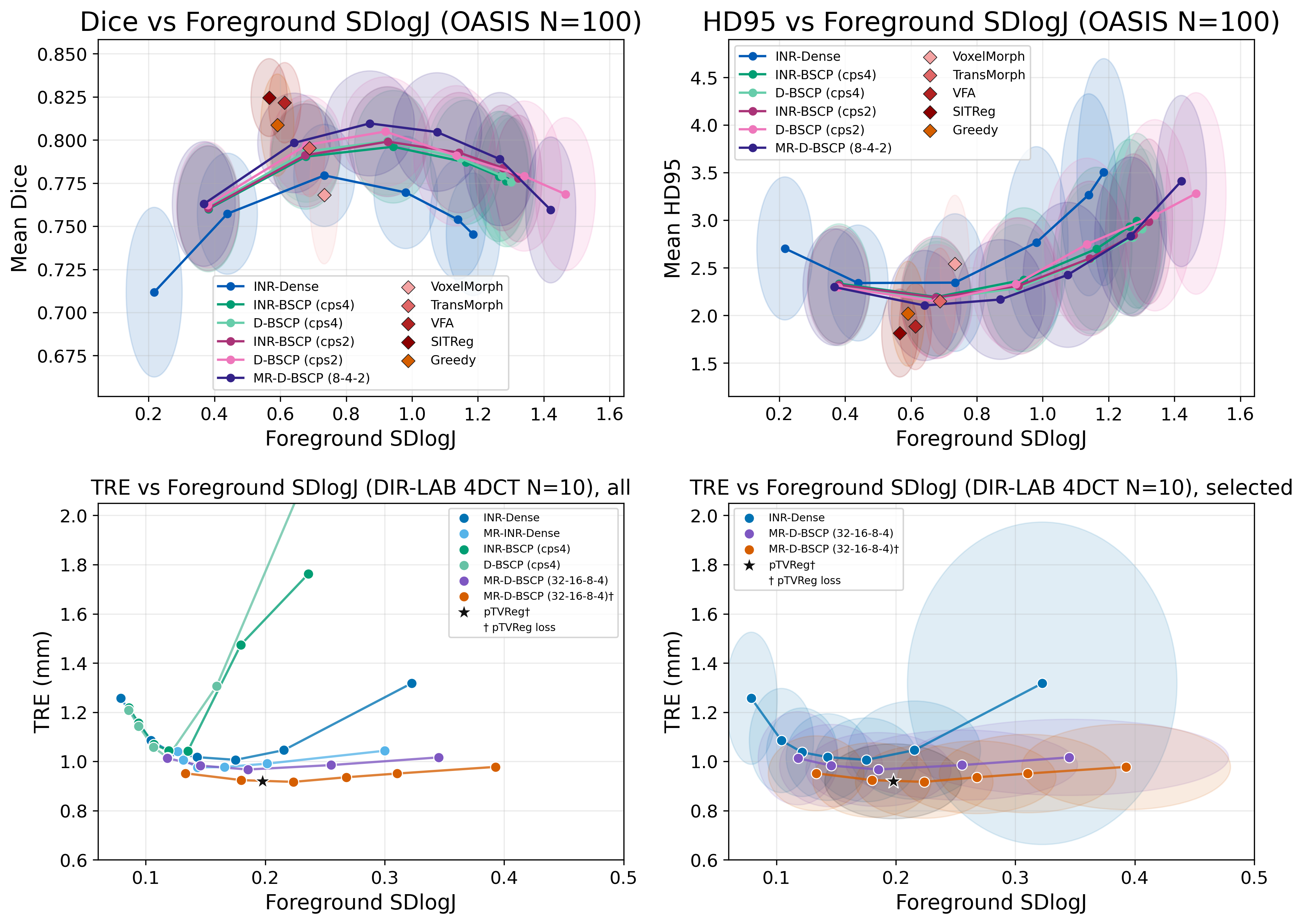}
\caption{Registration performance across the alignment-regularity spectrum on OASIS brain MR (top) and DIR-LAB 4DCT (bottom). Ellipse axes indicate standard deviation across cases.}
\label{fig:arc}
\end{figure}

Fig.~\ref{fig:arc} summarizes registration performance across the alignment-regularity spectrum. On OASIS, D-BSCP closely follows INR-BSCP across the Dice versus $\mathrm{SD}(\log J)$ and HD95 versus $\mathrm{SD}(\log J)$ curves at both 2 mm and 4 mm control-point spacing (cps). In several regions, D-BSCP performs slightly better. Thus, similar performance to INR-BSCP can be obtained by directly optimizing the B-Spline control points without the INR. MR-D-BSCP (8-4-2) achieves the best performance among the tested continuous parameterizations. In contrast, INR-Dense performs worse and occupies a smoother region of the trade-off, suggesting that its model prior may limit its ability to capture locally varying inter-subject brain deformations. Among external references, SITReg, VFA, and Greedy remain stronger operating points, while earlier learning-based methods without registration-specific design, especially multiresolution refinement, are less competitive~\cite{RethinkReg2024,RevisitDIR2025}.

DIR-LAB 4DCT shows a different behavior. Single-scale INR-BSCP and D-BSCP are less robust to large respiratory motion and can fail sharply at weaker regularization. INR-Dense is the most robust single-scale method on this task. MR-D-BSCP (32-16-8-4) substantially improves the spline-based formulation and closely matches the state-of-the-art pTVReg performance when using the same loss function as pTVReg ($0.917 \pm 0.148$ mm for MR-D-BSCP versus $0.919 \pm 0.152$ mm for pTVReg). These results suggest that large respiratory motion favors either a dense INR prior that promotes smooth, coherent deformation or, for local B-Spline control points, a coarse-to-fine optimization path that captures large displacement before local refinement.

MR-INR-Dense, adapted from Dual-INR~\cite{Dual-INR2026}, further supports the benefits of multiscale modeling for lung CT by decomposing motion into coarse and fine INR branches. It improves over INR-Dense, although it still does not match pTVReg's TRE. However, unlike B-Spline control lattices, INR parameters cannot be directly upsampled for finer-resolution refinement and instead require residual branches, composition, or continued optimization.

We also evaluated whether these pair-specific optimizers can be made faster without sacrificing the registration accuracy. We focus this runtime analysis on OASIS, where all registrations share the same image size; DIR-LAB cases differ in crop size and motion difficulty, making a compact runtime comparison less controlled. For each spline-based method, we kept the objective and regularization weight fixed and screened learning-rate multipliers, shortened optimization-step budgets, and adaptive early stopping on OASIS. The selected schedules were then evaluated on the OASIS-100 test set and timed from input images to the saved dense displacement field, including full optimization but excluding evaluation metrics. Table~\ref{tab:oasis_runtime} reports both
cold-process and warm-process runtimes, since the first case in a Python/CUDA process includes one-time library and kernel setup. The optimized schedules preserve OASIS-100 Dice closely while substantially reducing runtime; MR-D-BSCP in particular remains the most accurate tested continuous parameterization while reducing warm input-to-DVF time from about 62~s to 11~s.

\begin{table}[t]
\centering
\caption{Runtime-reduced OASIS schedules versus their 2500-step counterparts.
Accuracy is OASIS-100. Runtime is input-image to saved dense displacement field,
including optimization and DVF output, excluding metric evaluation.
Cold indicates a fresh process/CUDA context; warm indicates persistent-process
timing after warmup.}
\label{tab:oasis_runtime}
\fontsize{8}{9}\selectfont
\setlength{\tabcolsep}{2.5pt}
\begin{tabular}{@{}lccccc@{}}
\hline
Method & Dice & Steps & Cold (s) & Warm (s) & Peak GPU (GB) \\
 & base$\rightarrow$fast & base$\rightarrow$fast & base$\rightarrow$fast & base$\rightarrow$fast & - \\
\hline
MR-D-BSCP (8-4-2) &
0.810$\rightarrow$0.809 & 2500$\rightarrow$440 &
152.0$\rightarrow$101.1 & 62.4$\rightarrow$11.0 & 3.93 \\
D-BSCP (cps2) &
0.805$\rightarrow$0.805 & 2500$\rightarrow$431 &
135.4$\rightarrow$76.9 & 80.9$\rightarrow$14.4 & 2.16 \\
D-BSCP (cps4) &
0.799$\rightarrow$0.797 & 2500$\rightarrow$477 &
126.3$\rightarrow$78.6 & 80.9$\rightarrow$15.3 & 1.83 \\
INR-BSCP (cps2) &
0.799$\rightarrow$0.793 & 2500$\rightarrow$799 &
370.9$\rightarrow$171.4 & 277.9$\rightarrow$90.2 & 7.72 \\
INR-BSCP (cps4) &
0.796$\rightarrow$0.786 & 2500$\rightarrow$644 &
175.0$\rightarrow$90.4 & 107.8$\rightarrow$29.2 & 2.58 \\
\hline
\end{tabular}
\end{table}

\subsection{Target-field Fitting vs Registration Diagnostics}

\begin{figure}
\centering
\includegraphics[width=\textwidth]{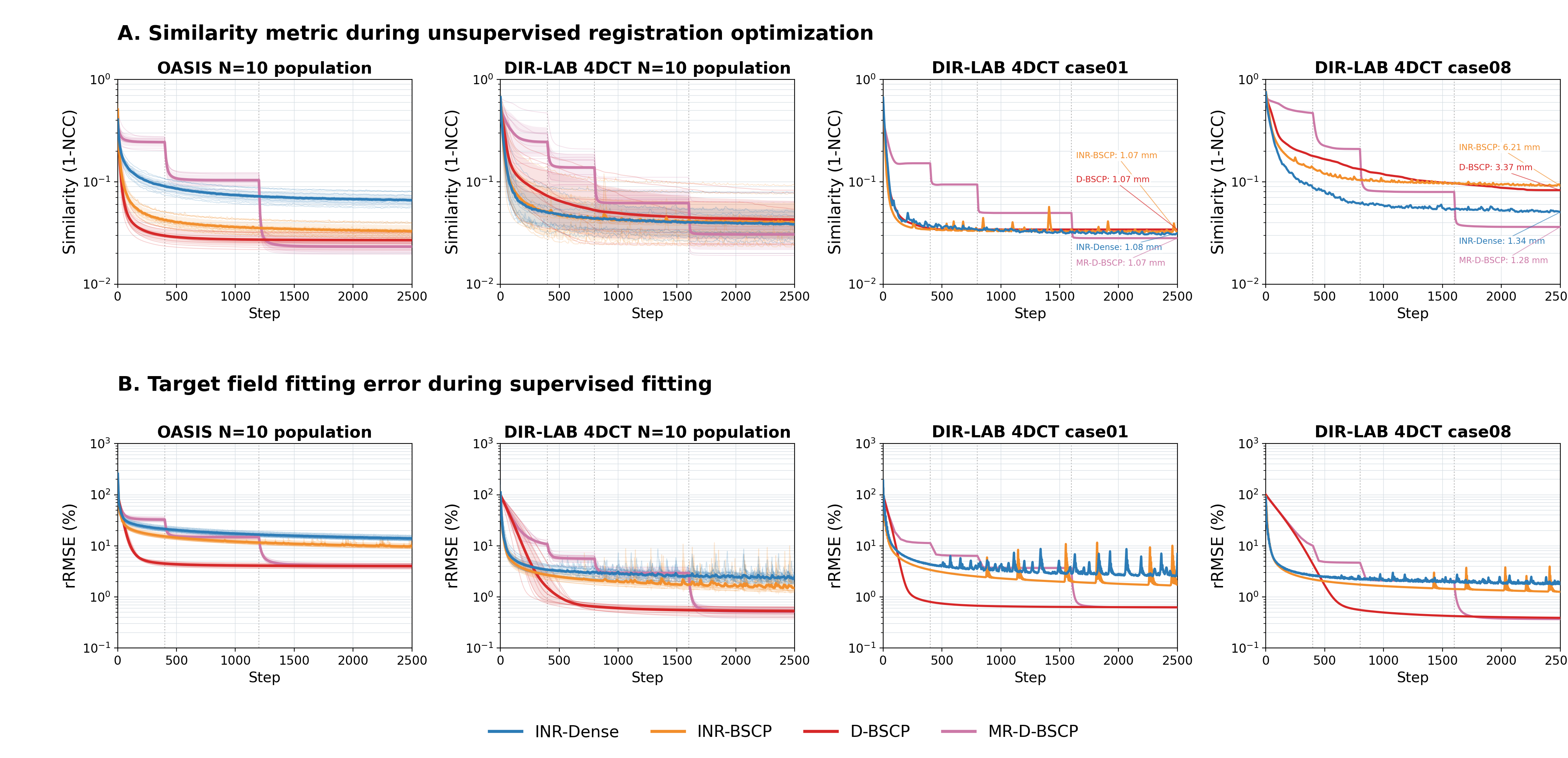}
\caption{Unsupervised registration versus supervised target-field fitting diagnostic for population and representative cases. In population panels, faint lines are individual cases, thick lines are log-space population means, and shaded regions show \(\pm 1\) log-space standard deviation.}
\label{fig:fitting}
\end{figure}

Fig.~\ref{fig:fitting} examines representation capacity versus registration optimization behavior by comparing supervised target-field fitting with unsupervised registration. On OASIS, the 10 sampled pairs show highly consistent behavior across methods, so we report only the population-level results. DIR-LAB is more heterogeneous. We therefore additionally show case01, where the methods behave similarly well, and case08, where single-scale methods struggle under large motion. 

Under supervised fitting, D-BSCP fits the reference fields well, with MR-D-BSCP achieving the lowest fitting errors. INR-BSCP and INR-Dense generally show higher errors. This pattern is consistent across the OASIS and DIR-LAB cases.

The contrast is most evident on the large-motion DIR-LAB case08. D-BSCP fits the target field well under supervised fitting, showing that the single-scale B-Spline model has sufficient representation capacity, yet it fails during unsupervised registration. INR-Dense shows the opposite trend. Its target-field fitting error is substantially higher, but it achieves better registration under large motion. When the model prior matches the deformation pattern, as in INR-Dense for smooth and coherent lung motion, good registration can still be achieved despite lower representation capability. Conversely, when the prior or optimization path is mismatched, high representation capacity alone may not prevent failure. In this case, the strong improvement from multiresolution optimization supports the interpretation that the single-scale methods fail because they converge to poor local minima. These results suggest that both the parameterization and the optimization scheme should be chosen according to the deformation properties of the target task.

\section{Conclusion and Discussion}
We presented a controlled validation study of continuous deformable registration parameterizations across inter-subject brain MR and exhale-to-inhale lung CT registration. Our results show that the success of an "off-grid" registration model is not determined by the use of a better continuous model, but by whether the induced deformation prior matches the target motion pattern. On OASIS, direct B-Spline control-point (D-BSCP) optimization closely matches INR-BSCP (SINR), suggesting that the B-Spline control-point prior explains much of INR-BSCP's favorable behavior. Moreover, our runtime-reduction study shows that D-BSCP and MR-D-BSCP can be made substantially faster on OASIS while preserving registration accuracy. On DIR-LAB, large respiratory motion requires long-range capture behavior, where INR-Dense (IDIR) and especially multiresolution methods are more suitable than single-scale B-Spline methods.

Our findings echo recent studies in deep learning-based DIR, which show that registration-specific designs matter more than naively adopting a new network architecture~\cite{RethinkReg2024,RevisitDIR2025}. Our study extends this message to continuous pair-specific registration: INR-based methods are useful, but they are not automatically superior to well-established B-Spline parameterizations when validated carefully. This motivates more rigorous validation of new registration methods, including ablation of the components that actually drive performance. We also highlight the importance of evaluating methods across the alignment-regularity spectrum~\cite{ARC2025},  rather than at a single operating point, to support fair comparison across different methods.

At the same time, our study does not isolate all sources of prior, particularly the effect of different regularization forms, and broader validation across registration tasks is still needed. More broadly, these results suggest that registration models should be designed around the deformation properties of the target anatomy. Similar ideas have been explored through sliding-motion, anatomy-adaptive, and biomechanics-informed priors~\cite{SlidingMotion2014,MUSA2025,BiomechPrior2023}. Developing task-specific deformation priors remains a promising direction for DIR~\cite{RegularizationReview2026}.

    

\begin{credits}
\subsubsection{\ackname} This study was funded by NIH R01EB031577.

\subsubsection{\discintname}
The authors have no competing interests to declare that are
relevant to the content of this article.
\end{credits}

%
%
%
%

\bibliographystyle{splncs04}
\bibliography{references_v2}





\end{document}